\documentclass{article}

\usepackage[preprint]{corl_2026} % Uncomment for pre-prints (e.g., arxiv); This is like ``final'', but will remove the CORL footnote.

\title{Inductive Generalization for Robotic Manipulation}

\author{
  Annabella Macaluso \quad
  Haochen Zhang \quad
  Ishaan Masilamony \\[0.4em]
  \textbf{Yingshan Chang} \quad
  \textbf{Yonatan Bisk} \\[0.4em]
  Carnegie Mellon University
}

\usepackage{times}

\usepackage[numbers]{natbib}
\usepackage{multicol}
\usepackage{graphicx}

\usepackage{bm}
\usepackage{xcolor}
\usepackage{multirow}
\usepackage{amssymb}
\usepackage{amsmath}
\usepackage{amsfonts}
\usepackage{amsthm}
\usepackage{booktabs}
\usepackage{wrapfig}
\usepackage{subcaption}
\usepackage[table]{xcolor}
\usepackage{colortbl}
\definecolor{annapurple}{RGB}{230, 210, 255} % 
\definecolor{palered}{RGB}{242, 178, 183}

\usepackage{longtable}
\usepackage{ragged2e}
\definecolor{lightgray}{gray}{0.93}
\newcolumntype{P}[1]{>{\RaggedRight\arraybackslash}p{#1}}

\newtheorem{assumption}{Assumption}[section]

\newcommand{\drop}[1]{\cellcolor{palered}\textbf{#1}}
\usepackage[most]{tcolorbox}
\usepackage{xcolor}

\definecolor{defblue}{RGB}{20,20,120}

\newtcolorbox[auto counter]{definitionbox}[1][]{
    colback=blue!5!white,
    colframe=defblue,
    fonttitle=\bfseries,
    title=Definition~\thetcbcounter: #1,
    boxrule=0.8pt,
    arc=4pt
}

\usepackage[size=tiny]{todonotes}

\begin{document}
\maketitle

\begin{abstract}
Understanding the generalization capabilities of visuomotor policies is essential in the development of capable robotic agents. Generalizable models learn structures that transfer across domains. However, in practice, visuomotor policies test performance by interpolation on known distributions using unstructured domain shifts (e.g. lighting, clutter, diverse objects). We argue that to measure generalization capabilities  we must instead test the inductive capacity of policies on progressively harder, out-of-distribution task variants. We call this \textit{inductive generalization}, drawing directly on how axis-based evaluation has revealed inherent generalization limitations in language models (e.g. sequence length, counting)~\cite{chang2025learningmodelsuccessors}. We provide a reusable and formal evaluation protocol for measuring inductive generalization in any manipulation policy, and establish baselines showing that existing paradigms fail this test; e.g. SoTA Vision-Language-Action models and find that policies that appear to generalize to prior domain shifts (distractors, etc) fail inductive generalization tests. These results expose a class of learning challenges orthogonal to those addressed by data and model scaling in robot learning, yet are imperative to solve in order to realize general purpose robots.
\end{abstract}

\section{Introduction}
\label{sec:introduction}

A central hypothesis of Large Visuomotor Policies, such as Vision Language Action models (VLAs) \cite{black2025pi, kim2024openvla, li2025dsvla, zhen20243d, hu2023toward, open_x_embodiment_rt_x_2023}, is that data scale paired with strong algorithms will give rise to generalizable policies. Directly inspired by the success of internet-scale pretraining in language and vision \cite{dosovitskiy2020image, comanici2025gemini}, the robotics community is investing heavily in large-scale data collection \cite{buildaiegocentric10k2025, open_x_embodiment_rt_x_2023, khazatsky2024droid} with the goal of achieving measurable progress on standardized benchmarks \cite{liu2023libero, robocasa2024} and real-world scenarios. As these general-purpose models grow in scale and capability, reliably evaluating their generalization capability becomes essential. Prior works have explored domain shifts via varying visual factors, such as the lighting conditions or object appearance \cite{secant_visual_perturb, darla_perceptual_domain_shift, tower_visual_appearance_shift, gen_rl_soft_dataaug}. However, these shifts often introduce qualitative differences in a task instead of quantitative measures. 

\begin{figure}[t]
    \centering
    \includegraphics[width=\linewidth]{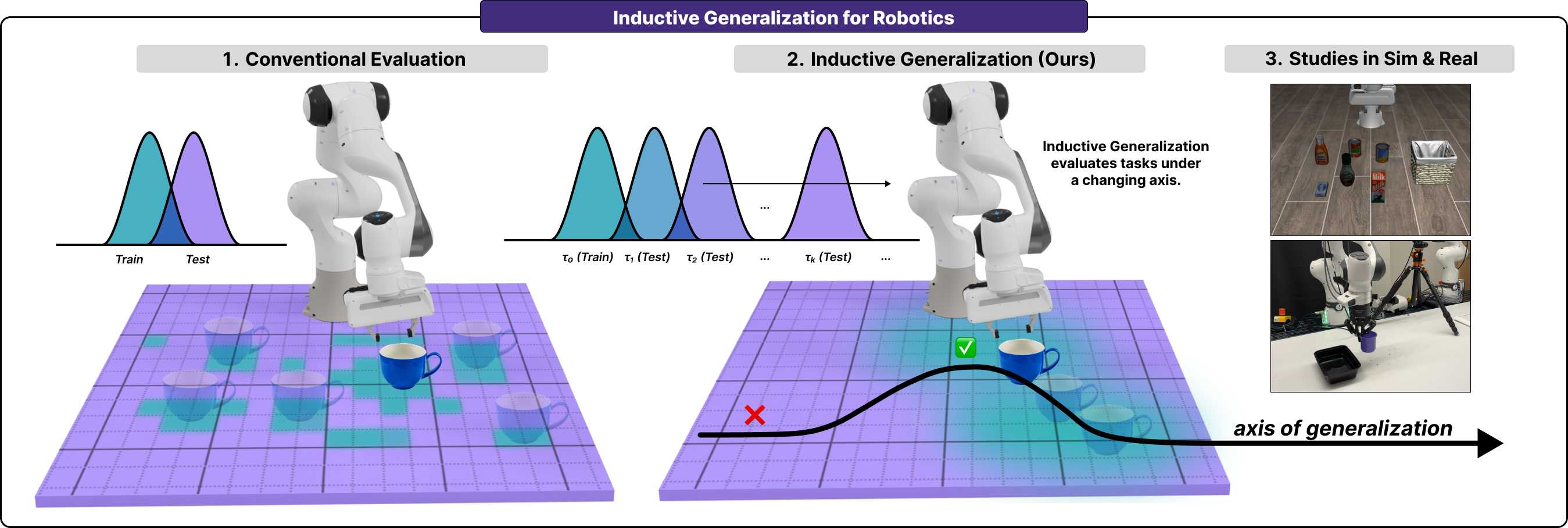}
    \caption{Where OOD generalization traditionally refers to testing on unseen \textit{samples}, these are drawn without regard to the underlying task structure, focusing instead on instance-level coverage (left). We contend a single train-test split does not admit the ability to distinguish generalization capability easily. Thus, this work focuses on Inductive Generalization, which tasks a model with inferring task invariances from a minimal set of samples that generalize to novel task $\tau$ instances.}
    \label{fig:teaser}
    \vspace{-2em}
\end{figure}

Addressing this requires first characterizing which problem classes resist simple solutions through interpolation or sheer training data scale. The Large Language Models (LLMs) community has surfaced several persistent generalization shortcomings, such as length generalization \cite{posenc_on_lengen}, counting \cite{chang2025language} or search over graphs \cite{transformer_search}. Most importantly, these limitations become apparent when models are tested along a well-defined axis of difficulty that progressively moves Out-of-Distribution (OOD) \cite{chang2025learningmodelsuccessors}. In robotics, while interaction with the environment is stochastic and challenging, large policies are performing better and better on in-distribution tasks \cite{black2025pi, bjorck2025gr00t}. Yet, it still has not been demonstrated whether these policies perform well on task spaces that evolve such as navigating through clutter to manipulate objects or executing tasks with increasingly long horizons. The solutions to these problem instances characterize inductive generalization: the capacity of a policy to extend a behavior it has learned at bounded task complexity (e.g. a pick behavior) to structurally identical problems of greater, unseen depth (picking many objects to reveal the goal object). We hypothesize that robotic manipulation tasks admit hidden \textit{inductive axes} along which extrapolation should be possible if a policy has discovered and captured the underlying axes or invariances. To this end, we provide a simple, yet powerful robotic task that contain underlying physical inductive axes, but are not solvable by scale and data alone. Instead, realizing the underlying inductive axes and achieving \textit{inductive generalization} (Figure \ref{fig:teaser}) is necessary in order to solve such problems. 

A complementary viewpoint is one that's grounded in real robot problems. Robotic manipulation evaluation through the lens of \textit{inductive generalization}, highlights whether models capture task-solving rules as tasks advance along well-defined axes of difficulty beyond their training support. By framing evaluation around explicit inductive axes, we distinguish policies that capture the true invariance from those that rely solely on broad data coverage. For instance, in block-stacking, enlarging the training coverage provides only a provisional remedy, because it does not equip the robot with the ability to anticipate future expansions beyond the training coverage. 

Our experiments show that all evaluated SoTA VLAs achieve strong in-distribution performance yet fail systematically along structured spatial and compositional axes. We find supplementing training with additional in-distribution data (via MimicGen \cite{mandlekar2023mimicgen}) provides no improvement. This indicates that failures are not addressable by data coverage. This is not to say data coverage is without value, instead this implies inductive generalization is a separate capability that does not emerge as a byproduct of a scale. We provide a formal evaluation framework for measuring this capability in any manipulation policy, and establish baselines documenting the gap. We view this as a necessary foundation for the architectural, training, and data-design work that must follow.
\section{Related Work}
\label{sec:relatedwork}
\vspace{-0.75em}
\textbf{Observations from LLM Learning.} The LLM learning community actively searches for persistent generalization failures despite large-scale training \cite{NEURIPS2020_1457c0d6, finzi2026entropy, lotfi2023non}. A key observation is that models with near-identical pretraining loss can exhibit substantially different behavior on generalization tests, motivating evaluation in the \textit{saturation regime} \cite{reizinger2024position, pmlr-v202-liu23ao}. Puzzles revealed by saturation regime evaluations have sparked discussions on understanding and injecting inductive biases into the model architecture \cite{chang2025language, math_inductive_bias_lm, inductive_bias_looped_transformers, NAR}, and on exploring alternative architecture families \cite{mamba, RWKV}. Transferring insights from language modeling to robot learning, it is similarly crucial to pay attention to saturation regimes and the inductive biases that enable generalizable behavior. 

\vspace{0.5em}
\textbf{Methods for Generalization in Robotics.} A prominent direction for developing \textit{generalist policies} is learning multi-task and hierarchical policies \cite{yu2020meta, finn2017model, hierarchical_rl_gen_subgraph, hierarchical_rl_manipulation}, where policies are trained across a distribution of tasks that have shared underlying representations. Such methods rely on the assumption that unseen evaluation tasks are still drawn from the same distribution as training. Integrating LLMs and Vision Language Models (VLMs) \cite{ahn2022can, driess2023palm, kang2024clip, pan2025omnimanip, liu2025robodexvlm} into modular robotics pipelines unlocked unprecedented zero-shot reasoning capacity. Recently, vision-language-action (VLA) models that combine a pretrained VLM backbone with an action head \cite{black2025pi, clark2025action} lead to further performance boosts. There is an implicit assumption that ``open-world generalization'' can emerge by virtue of large data coverage and pretrained vision-language priors. However, it is unclear what is actually in-distribution (ID) or out-of-distribution for VLA models or how much data coverage is ``large enough". Note that LLM researchers continue to collect and annotate data to address coverage gaps, while VLAs face an even harsher data bottleneck due to their complex modalities. If data coverage drives generalization for models, it begs the question of why this is not true for humans, who can rapidly uncover patterns and extrapolate with far less data. Based on the belief that rigorous evaluation drives rigorous development, our core contribution is a generalization framework with a formalization of \textit{structured domain shift}, emphasizing the characterization of an inductive axis as a core technical component for designing evaluation suites.  

\vspace{0.5em}
\textbf{Robotic Manipulation Evaluation.} Existing robotic benchmarks \cite{mittal2025isaaclab, Xiang_2020_SAPIEN, todorov2012mujoco}, including RLBench \cite{9001253}, Maniskill \cite{tao2024maniskill3}, LIBERO \cite{liu2023libero} and robomimic \cite{mandlekar2021matters}, generally define generalization as success on \textit{unseen, individual} task instances using few-shot learning or in-distribution single task success. We provide further details of benchmark evaluation methods in Appendix \ref{app:benchmark}. While sufficient at measuring success over individual tasks, these methods conflate task performance with generalization capability. A policy that has merely memorized its training distribution is indistinguishable under these metrics. STAR-Gen \cite{gao2025taxonomy} recognized this gap, introducing a taxonomy that creates visual, behavioral or semantic categories for distribution shift. However, this method does not address identifying where axes lie within existing training distributions and their separation from test-time evaluation. Our work directly addresses this by introducing inductive difficulty progressions to measure this. Data generation techniques such as DemoGen \cite{xue2025demogen} seek structure-aware data augmentation, which is complementary, however, we are seeking structure-aware task design and evaluation. Without this it’s hard to even reveal the advantage of data augmentation methods.
\vspace{-0.25em}
\section{Setup}
\label{sec:setup}
\vspace{-0.75em}

This section standardizes the notation, current robot evaluation framework, and outlines the assumptions that inform our design choices.

\textbf{Notation}. Following the notation used by \citet{gao2025taxonomy}, we address robot manipulation policy-learning in continuous action spaces. We consider an environment defined by the tuple $E = (\mathcal{S}, \mathcal{O}, \mathcal{A}, \mathcal{G}, p_o, p_t, r)$. Here, the state space $\mathcal{S}$ and the action space $\mathcal{A}$ are continuous. Observations $o \in \mathcal{O}$ are generated from  states $s \in \mathcal{S}$ via a conditional distribution $p_o(o \mid s)$. An action $a \in \mathcal{A}$ is applied by the robot, and the unknown transition dynamics are described by $p_t(s' \mid s, a)$. The environment provides a bounded reward signal $r : \mathcal{S} \times \mathcal{A} \rightarrow [0,1]$ at each transition. We define the space of tasks $\mathcal{T}$ for an environment $E$ as $\mathcal{T} = (p_\tau(s_0), g_\tau, r_\tau)$. Here $p_\tau(s_0)$ is an initial state distribution of $E$. The goal $g_\tau$ is sampled from the space of possible goals, $\mathcal{G}$ within $E$, and $r_\tau$ is the reward signal of the goal $g_\tau$.

\vspace{-0.25em}

\subsection{Generalization Overview}

In evaluating robotic manipulation, policies operate over continuous state, action and goal spaces. To make robotic learning tractable, existing approaches introduce behavior-level abstractions (e.g. pick-and-place, stacking, insertion). However, even within a single behavior class, variations in object pose, shape, size, contacts and environment context induce an unbounded set of distinct interaction scenarios. Moreover, achieving robustness through naive data coverage is intractable because the observation-action input distribution and space of realizable tasks are defined over an unbounded space and under deployment. Therefore, the robot is likely to encounter inputs that have vanishingly low support in the training data. In order to overcome these limitations, learning transferable structure from robotic data which enables extrapolation beyond training support is especially crucial in this domain. In robot learning, this process is characterized as generalization, which allows policies to operate across different domains.

\begin{figure}[t]
    \centering
    \includegraphics[width=\linewidth]{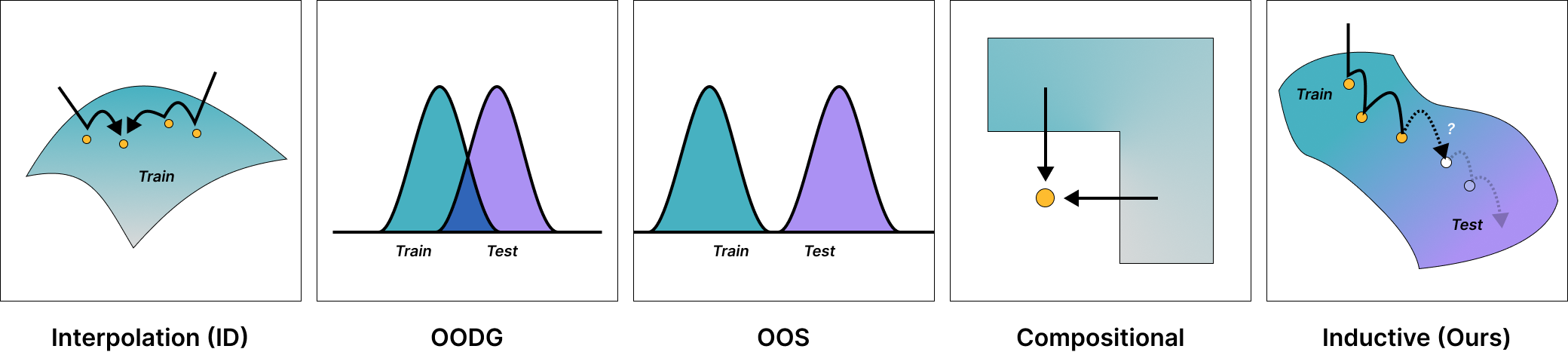}
    \caption{Common classifications include Interpolation within In-Distribution, Out-of-Distribution Generalization (OODG), Out-of-Support (OOS) and Compositional Generalization. Our method introduces inductive generalization for robotic manipulation where a single semantics-preserving axis defines a difficulty progression that moves tasks progressively out-of-distribution, distinguishing policies that capture true task invariances from those that merely interpolate.}
    \label{fig:ood}
    \vspace{-1em}
\end{figure}

The easiest setting to evaluate generalization is \textbf{In-Distribution Generalization} (IDG). IDG assumes that the test dataset $\mathcal{D}_{\text{test}}$ is drawn from the same distribution as the training dataset $\mathcal{D}_{\text{train}}$; i.e. $\mathcal{D}_{test} = \mathcal{D}_{train}$. This setting underlies several standard benchmarks for learning-based manipulation \cite{liu2023libero}. However, because continuous state and action spaces induce an effectively unbounded task landscape,  
distributional shifts in robot learning can easily break the i.i.d. assumption and make the dense sampling required for effective interpolation impractical. Thus, \textbf{Out-of-Distribution Generalization} (OODG) is motivated. However, it is possible to construct training and testing distributions that differ, but still share support. Finally, extrapolation demands an even harder scenario where they do not share support. That is, \textbf{OODG with Domain Shifts}.

Domain shift in robotics arises when the input distribution $p_\tau(s\mid s_0, F)$ changes due to variation in an external domain variable \(F\) \cite{netanyahu_methods_2025}. Domain shifts defined at an intuitive level often conflate multiple sources of variation. For example, introducing distractors simultaneously perturb an indeterminate set of pixels and state trajectory positions. To enable rigorous evaluation of robotic manipulation, it is necessary to isolate a single variation along a single axis at a time. Accordingly, we formalize \textbf{OODG with structured domain shifts} (in Section \ref{sec:method}), where a single, well-delineated inductive axis specifies the direction along which the task moves out-of-support \footnote{Our advocacy for structured domain shifts is orthogonal to curriculum learning \cite{bengio_CL}. Curriculum learning is a \textit{training} strategy that organizes learning into staged progressions, whereas our contribution is an \textit{evaluation} framework whose test suites captures a progression of difficulty in OODG, without \textit{presuming} or \textit{enforcing any particular learning procedure.}}. This formulation enables a transition from coarse generalization claims to a precise characterization of when and how policies transfer.

\vspace{-0.5em}
\section{Inductive Generalization}
\label{sec:method}
\vspace{-0.75em}
The current methodology of OODG suffers from two structural limitations. First, methods may fail to induce domain shifts where test and training distributions display reduced support. Thus, while strong performance can be achieved on proposed domain shifts it doesn't indicate the generalizability of the given method. Second, when OODG is evaluated, the task space is typically partitioned into binary in-domain and out-of-domain regions \cite{darla_perceptual_domain_shift}. This binary partitioning obscures how performance degrades as task difficulty progressively diverges from what can be revealed by training. Unlike conventional OOD evaluation, we are interested in whether a policy can extract rules and structures that allow it to maintain performance along a difficulty progression. Following Chang and Bisk \cite{chang2025learningmodelsuccessors}, one such method to invoke this capability is through a family of inductive problems.  

\begin{definitionbox}[Inductive Difficulty Progression, label=def:structured_difficulty]
% \label{def:structured_difficulty}
Let $\mathcal{T}$ be the space of robotic manipulation tasks. A structured \emph{inductive difficulty progression} consists of a base task distribution $\mathcal{T}_0\subset\mathcal{T}$ and a \emph{task successor} $f:\mathcal{T}\to\mathcal{T}$ generating a sequence
\[
    \mathcal{T}_{k+1} \;=\; f(\mathcal{T}_k), \qquad k = 0, 1, \ldots, K-1,
\]
which satisfies:
\begin{enumerate}
    \item \emph{Semantics preservation:} all tasks share a fixed success condition $\mathcal{R}$, so the goal of the task is unchanged across levels.\protect\footnote{Semantics-preserving means the progression does not alter the success condition $\mathcal{R}$ or the meaning of the task.}
    \item \emph{Strictly harder regime:} each $\mathcal{T}_{k+1}$ contains tasks outside the support of $\mathcal{T}_0,\ldots,\mathcal{T}_k$, varying a single, human-interpretable axis.
    \item \emph{Single-axis control:} $f$ varies only the axis of interest.
\end{enumerate}
\end{definitionbox}

\begin{assumption}[Atomic Step]
\label{def:assumption1}
Each transition $\mathcal{T}_k\to\mathcal{T}_{k+1}$ corresponds to a single application of $\phi_{\mathrm{step}}$. Difficulty increases atomically along the axis.
\end{assumption}

\begin{assumption}[Progression Feasibility]
\label{def:assumption2}
We make the assumption that tasks along the progression are feasible and atomic, ie. involves a single change along the progression.    
\end{assumption}

\begin{definitionbox}[Inductive Generalization, label=def:inductive_gen]
% \label{def:inductive_gen}
Let $S(\pi,\mathcal{T}_k)$ denote the success rate of policy $\pi$ on $\mathcal{T}_k$. Given a threshold $\gamma\in (0,1]$, $\pi$ exhibits \emph{inductive generalization} on the progression $\{\mathcal{T}_k\}_{k=0}^{K}$ if
\[
    S(\pi,\mathcal{T}_k) \;\ge\; \gamma \qquad \text{for all } k\in\{0,\ldots,K\}.
\]
We summarize each policy by its \emph{break level}
\[
    k^\star(\pi) \;=\; \min\{k \in \{0, \ldots, K\} : S(\pi,\mathcal{T}_k) < \gamma\},
\]
with $k^\star(\pi)=K{+}1$ if no break occurs in the evaluated range.
\end{definitionbox}

In some problem spaces $k$ may be countably infinite, however for feasibility we introduce a bound $K$. We note that failure on $\mathcal{T}_{n+1}$ after training only on $\mathcal{T}_0$ does not refute inductive capacity as revealing the underlying progression may require training across $\mathcal{T}_0,\mathcal{T}_1,\ldots,\mathcal{T}_n$ for some $n<k^\star$, mirroring the multi-domain training signal required by inductive learners in~\cite{chang2025learningmodelsuccessors}. Conversely, robustness to unstructured domain shifts~\cite{tobin2017domain, darla_perceptual_domain_shift} does not imply inductive generalization as a policy may attain high $S$ under appearance shifts for a select $k$, yet have a small $k^\star$ along a semantics-preserving axis. While inductive biases can be incorporated a priori (some latent rules correspond to physical or geometric invariances), our goal is to evaluate whether policies can acquire the underlying progression from experience.

\section{Experiments}
\label{sec:experiments}
\vspace{-0.75em}

\begin{figure*}[ht]
\centering
\begin{subfigure}{0.24\textwidth}
  \centering
  \includegraphics[width=\linewidth]{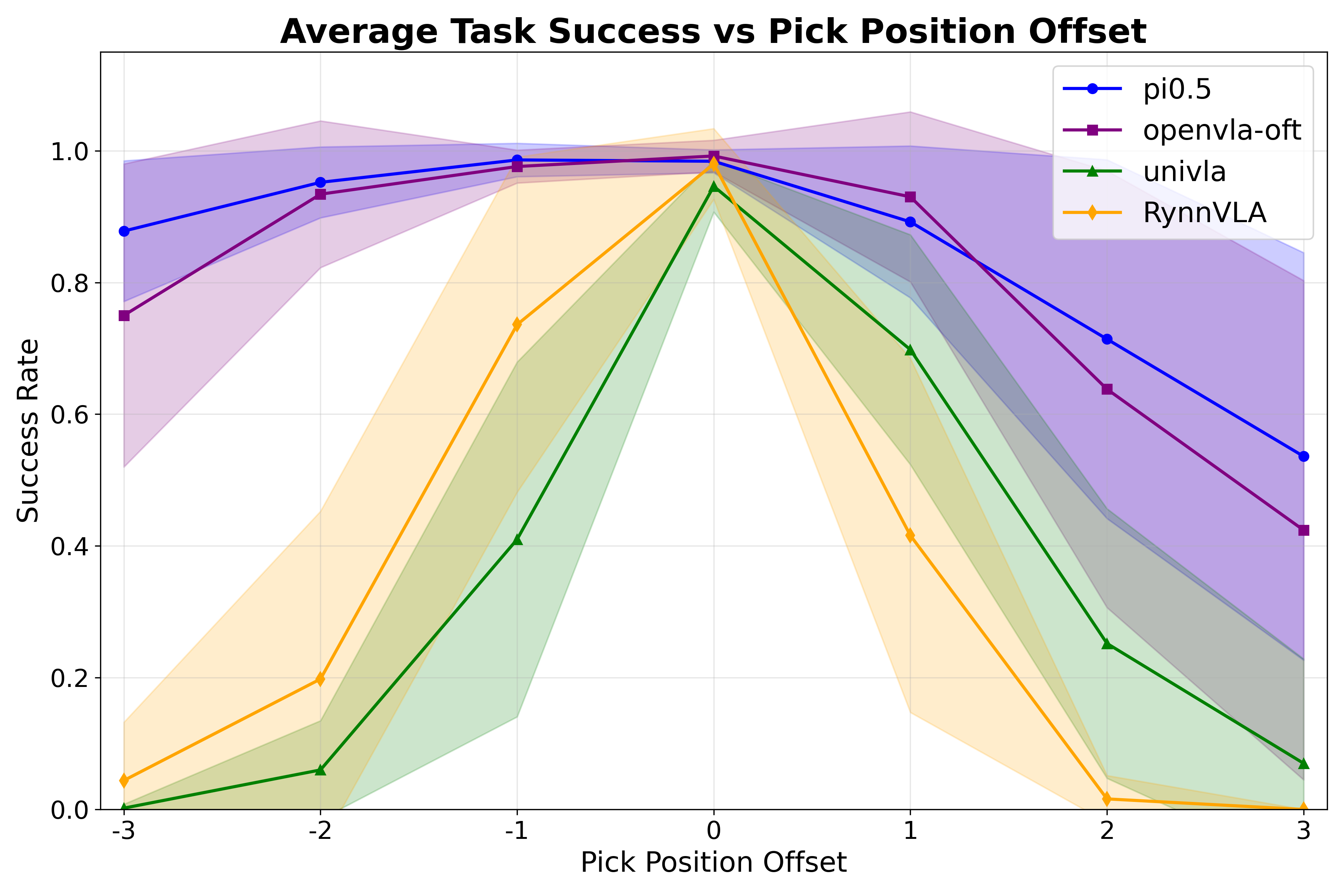}
\end{subfigure}
\hfill
\begin{subfigure}{0.24\textwidth}
  \centering
  \includegraphics[width=\linewidth]{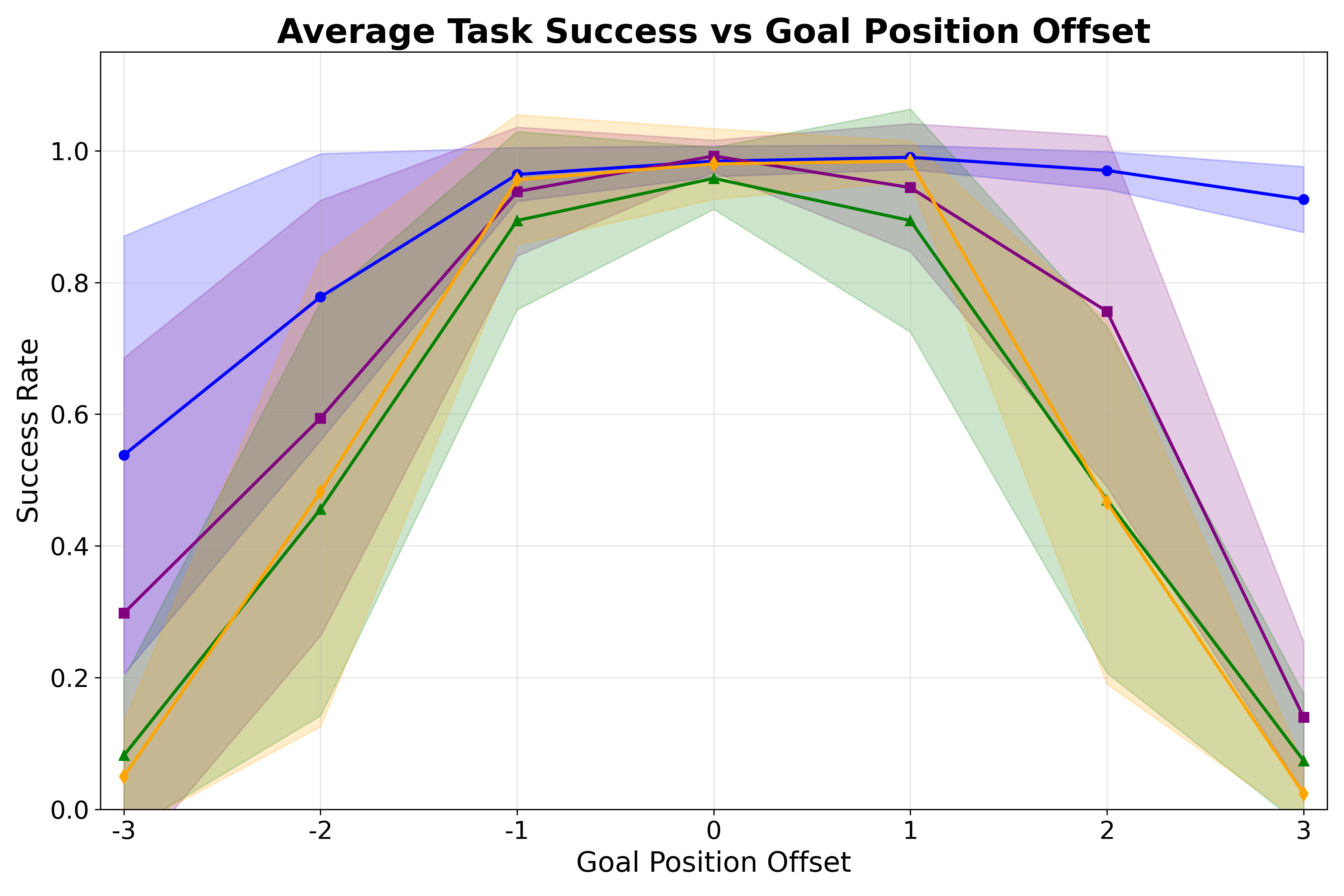}
\end{subfigure}
\hfill
\begin{subfigure}{0.24\textwidth}
    \centering
    \includegraphics[width=\linewidth]{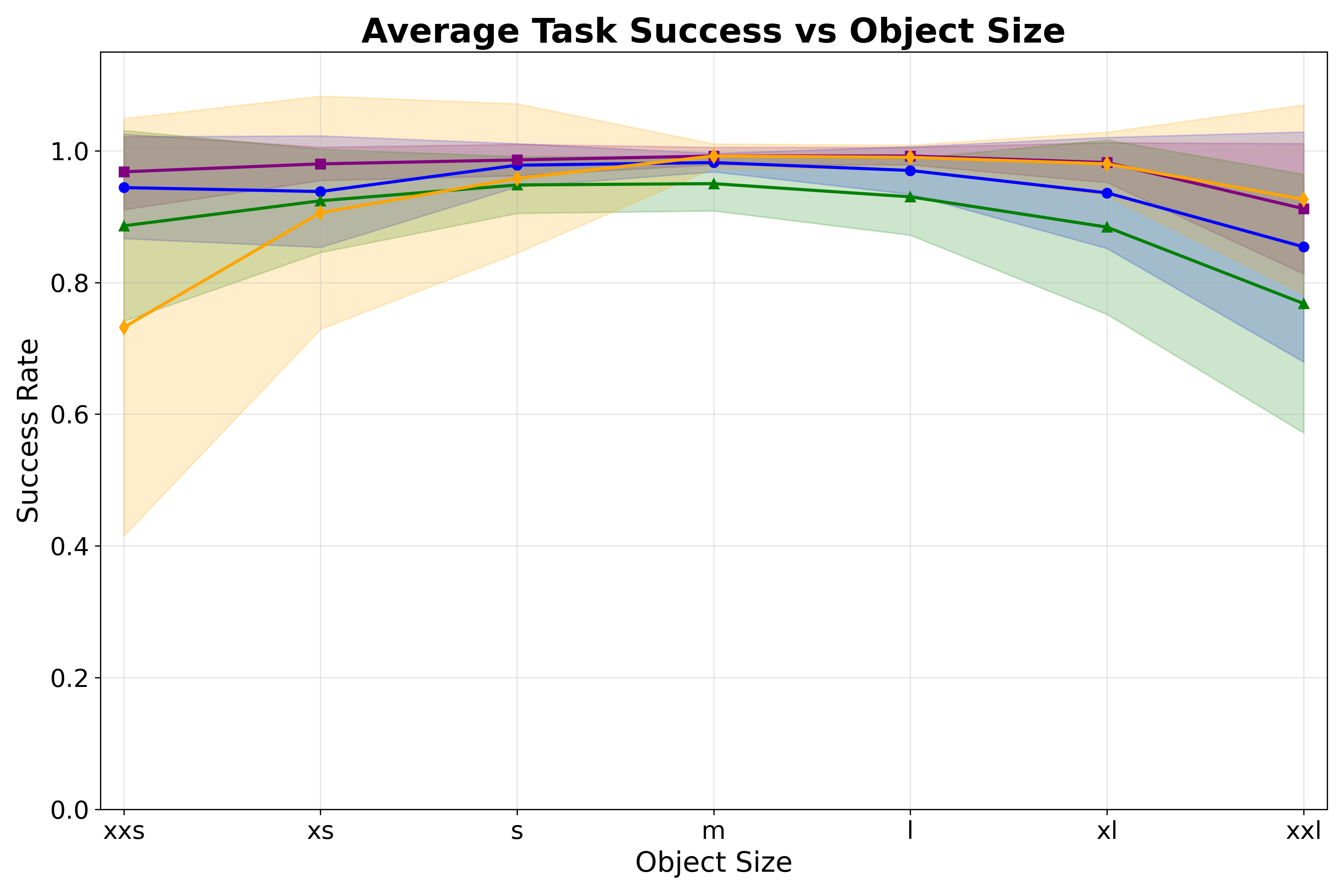}
\end{subfigure}
\hfill
\begin{subfigure}{0.24\textwidth}
    \centering
    \includegraphics[width=\linewidth]{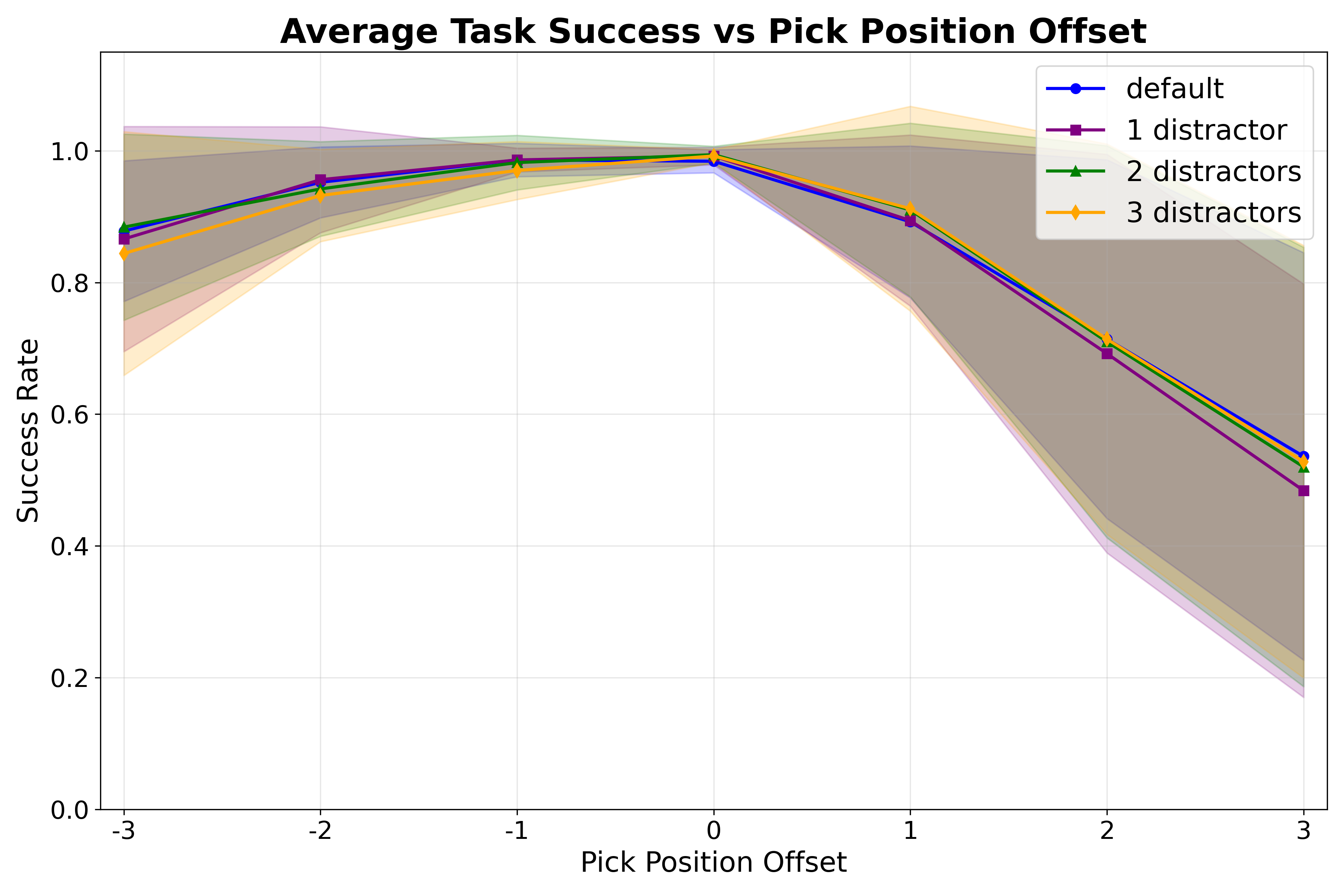}
\end{subfigure}
\caption{Results on probing VLAs along 4 separate axes: (1) Transformed pick location (2) Transformed goal location (3) Scale and (4) Composition of Pick and Visual Distractor on Pi0.5 only. The center of the graph, `0' demarcation, represents ID task used in training. Shifting left and right on the graph moves the task progressively OOD. }

\label{fig:libero_results}
\end{figure*}

\begin{figure*}[h]
    \centering

    \begin{subfigure}[b]{0.24\textwidth}
        \centering
        \includegraphics[width=\linewidth]{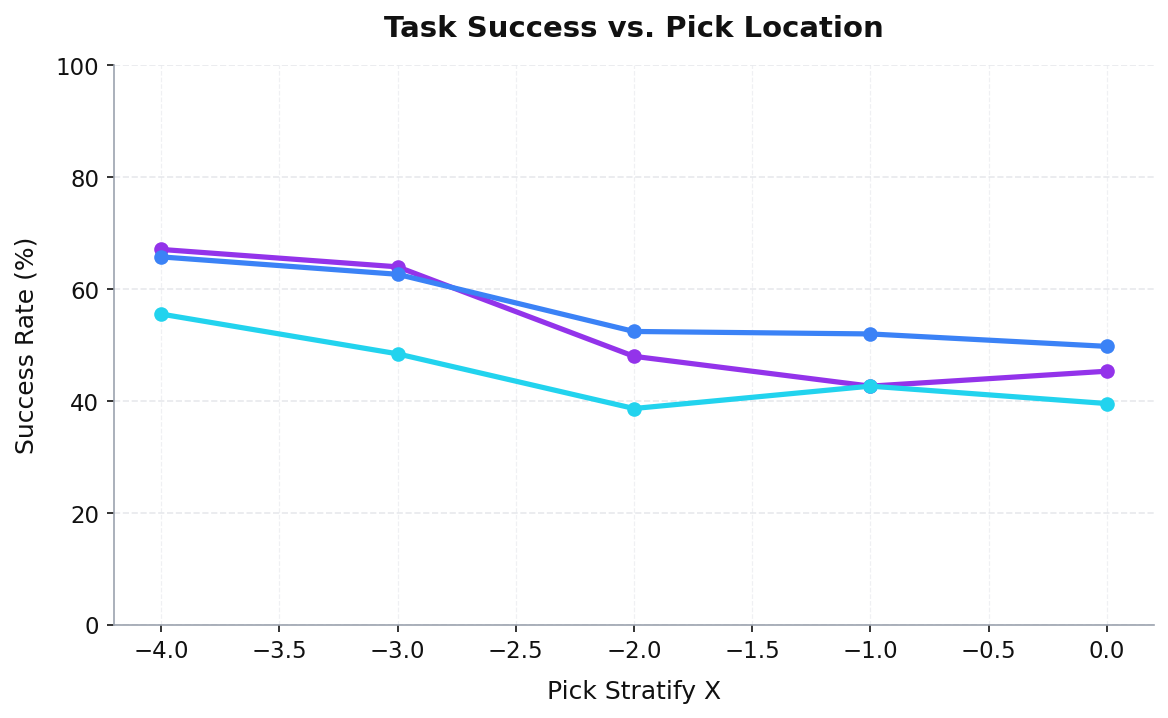}
        \label{fig:sub1}
    \end{subfigure}
    \hfill
    \begin{subfigure}[b]{0.24\textwidth}
        \centering
        \includegraphics[width=\linewidth]{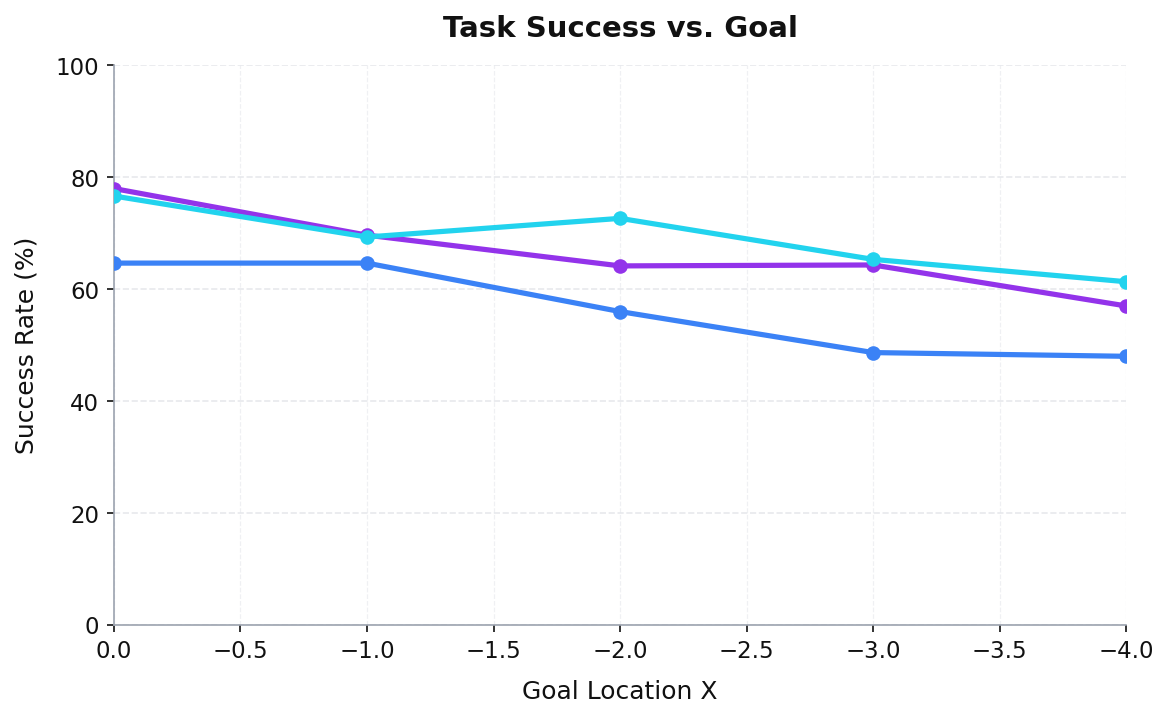}
        \label{fig:sub2}
    \end{subfigure}
    \hfill
    \begin{subfigure}[b]{0.24\textwidth}
        \centering
        \includegraphics[width=\linewidth]{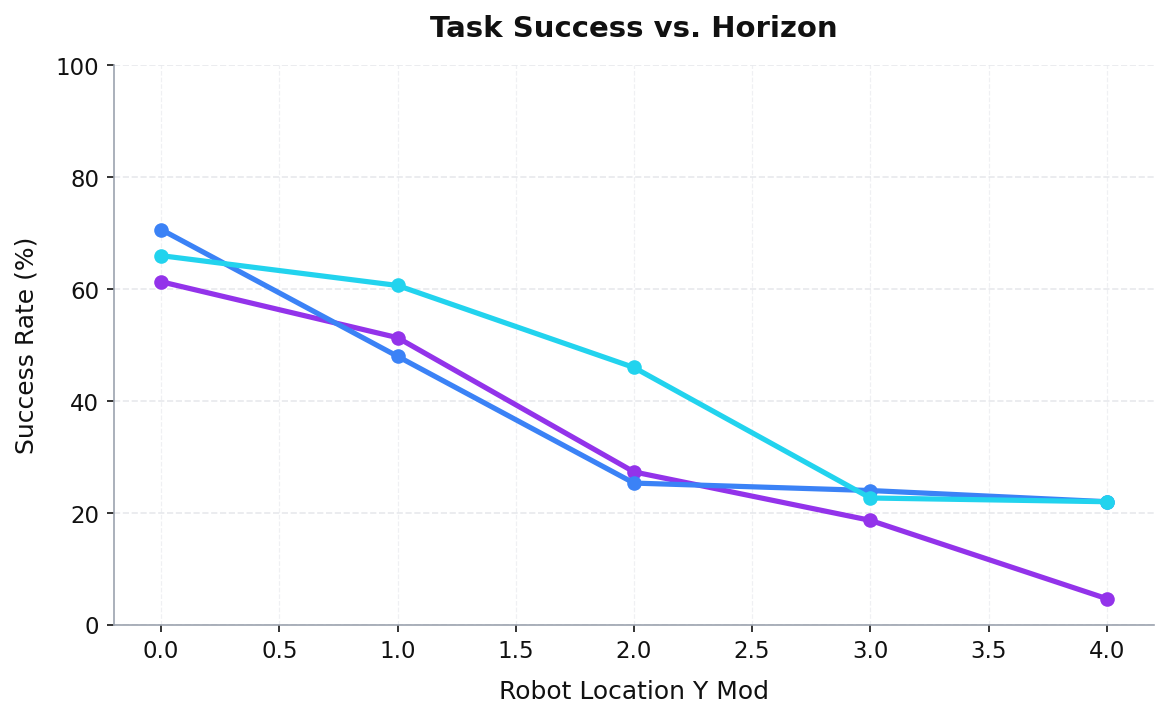}
        \label{fig:sub3}
    \end{subfigure}
    \hfill
    \begin{subfigure}[b]{0.24\textwidth}
        \centering
        \includegraphics[width=\linewidth]{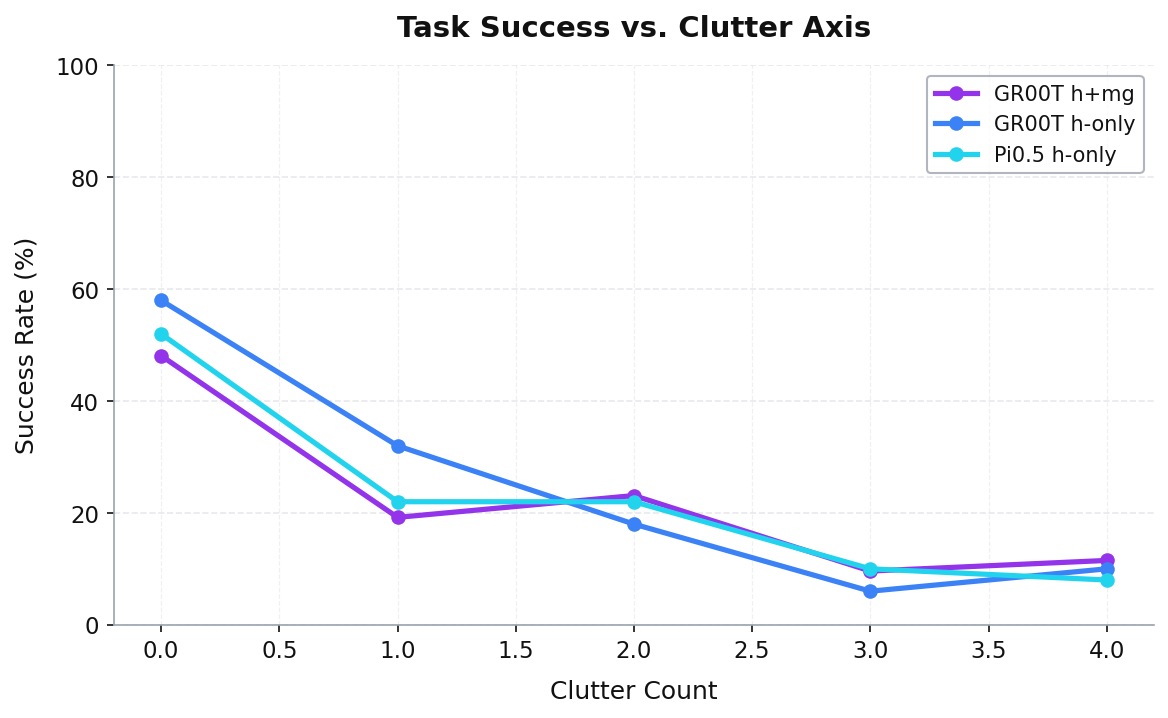}
        \label{fig:sub4}
    \end{subfigure}

    \caption{Inductive Generalization Study on RoboCasa Benchmark. Baselines used are Isaac-GR00T \cite{bjorck2025gr00t} and $\pi_{0.5}$\cite{black2025pi}. \textit{h-only} means only human demonstrations were used in policy training while \textit{h+mg} includes additional ID data created by MimicGen\cite{mandlekar2023mimicgen} in finetuning.}
    \label{fig:robocasaoverall}
\end{figure*}

In this section, we conduct experiments to answer the following questions: (i) How do we implement inductive difficulty progressions? (ii) How do current learning paradigms generalize? (iii) How does scaling data for visuomotor policies impact their inductive generalizability? (iv) How can we construct axes for testing in the real world? 

\vspace{-0.5em}
\subsection{Inductive Difficulty Progressions}
\vspace{-0.5em}

To provide an example of the inductive structure, we design a toy environment where the ground-truth inductive rule is known by construction. Illustrated in Figure~\ref{fig:blockenvillustration}, a parallel jaw gripper is tasked to grasp the yellow goal block. However, teal distractors block feasible grasps as two adjacent spots must be free in order to qualify as a valid grasp. Thus, teal blocks must be removed first in order to reveal a valid grasp. Each level induces a harder scenario and lets us observe whether learned policies discover the recursive rule to solve the task. We discuss more about the problem setting in depth in Appendix \ref{app:blockenv}. 

\definecolor{ood}{HTML}{D8CBFF}

\vspace{-0.5em}
\begin{table}[h]
\centering
\caption{Success rate (\%) on BlockEnv across inductive levels.
  Models are trained on L0--L2 only (shaded columns are
  OOD). \emph{Breaks at} denotes the first level
  where success drops below 20\%.}
\label{tab:block_env_results}
\begin{minipage}{0.55\linewidth}
  \centering
  \setlength{\tabcolsep}{4pt}
  \small
  \begin{tabular}{lcccc|c}
  \toprule
   & \multicolumn{3}{c|}{Success rate (\%)} & & \\
  \cmidrule(lr){2-5}
  Method
    & L0 & L1 & L2
    & \cellcolor{ood}L3
    & \multicolumn{1}{c}{Breaks at} \\
  \midrule
  MAML~\cite{finn2017model}
    & 100 & 100 & 100 & \cellcolor{ood} 0.0 & L3 \\
  RNN
    & 96.0 & 90.0 & 94.0 & \cellcolor{ood} 0.0  & L3 \\
  Transformer
    & 97.0 & 79.0 & 93.0 & \cellcolor{ood} 4.0 & L3 \\
  Transformer+RoPE
    & 97.0 & 88.0 & 90.0 & \cellcolor{ood} 1.0 & L3 \\
  Transformer+Goal/Act
    & 96.0 & 81.0 & 88.0 & \cellcolor{ood}0.0  & L3 \\
  TRM~\cite{jolicoeur2025less}
    & 100 & 100 & 92.0 & \cellcolor{ood}0.0  & L3 \\
  \textbf{Ind (ours)}
    & \textbf{100} & \textbf{100} & \textbf{100}
    & \cellcolor{ood}\textbf{86.0} & \textbf{L5} \\
  \midrule
  Oracle & 100 & 100 & 100 & \cellcolor{ood}100  & -- \\
  \bottomrule
  \end{tabular}
  \label{tab:blockenv}
\end{minipage}
\hfill
\begin{minipage}{0.35\linewidth}
  \centering
  \includegraphics[width=\linewidth]{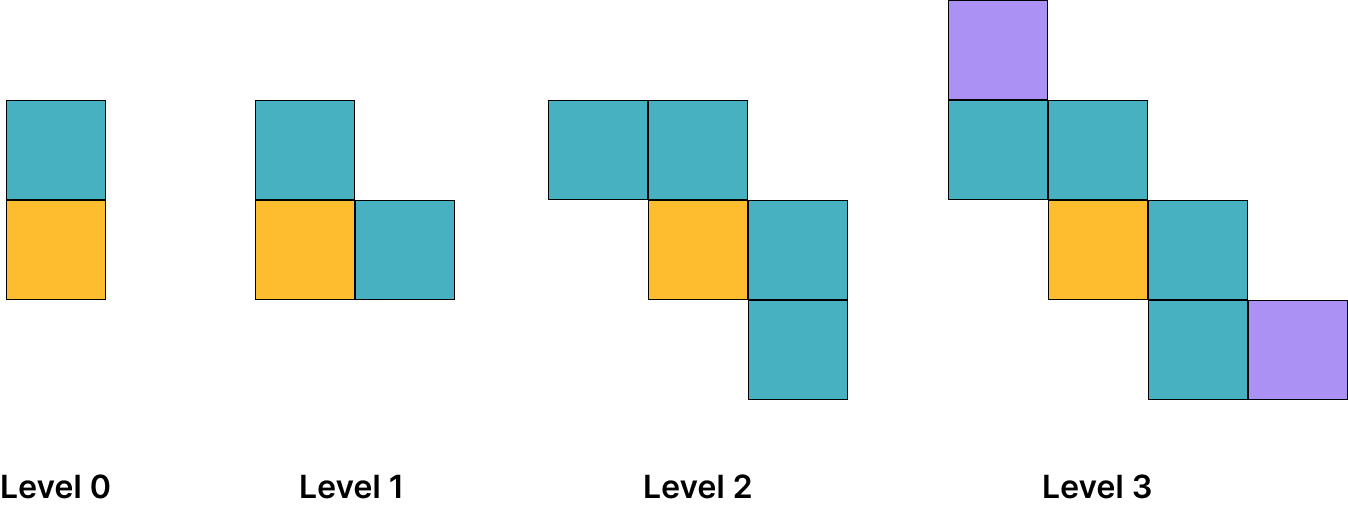}
  \vspace{1em}
  \captionof{figure}{Toy block env. Yellow denotes the goal block a gripper should grasp, but teal blocks must be removed first in order to achieve a valid grasp.}
  \label{fig:blockenvillustration}
\end{minipage}
\end{table}

\textbf{How do current learning paradigms generalize?} We evaluated policy classes spanning from purely data-driven to explicitly structured in Table \ref{tab:blockenv}. We construct a method, Ind, that augments behavior cloning with an auxiliary grasp prediction head, which learns the local invariant grasping rule (Appendix~\ref{app:blockenv}). \textbf{Takeaways.} (i) \emph{In-distribution competence does not imply inductive generalization}: high success rates through L2 are compatible with complete OOD failure, as seen for Transformers and TRM~\cite{jolicoeur2025less}. (ii)~\emph{Modeling explicit inductive structure, Ind,
extends but does not resolve the generalization gap}: encoding the local grasping
invariant pushes the break level from L3 to L5, yet eventual failure confirms that the learned rule was incomplete.

\subsection{Inductive Generalization on Standard Benchmarks}  
 Having established a controlled setting, a question that remains is whether data and large-scale pre-training reveals inductive generalization capabilities. To test for this, we evaluate SoTA VLA's on constructed inductive axes in two benchmarks: (1) LIBERO a standard, controlled benchmark which SoTA VLAs test against and (2) RoboCasa \cite{nasiriany2026robocasa365} which contains 65 atomic manipulation tasks in photorealistic kitchen environments with over 2,500 kitchen environments and object assets; providing scale and diversity for testing inductive axes. 

\begin{figure*}[t]
    \centering
    \includegraphics[width=\linewidth]{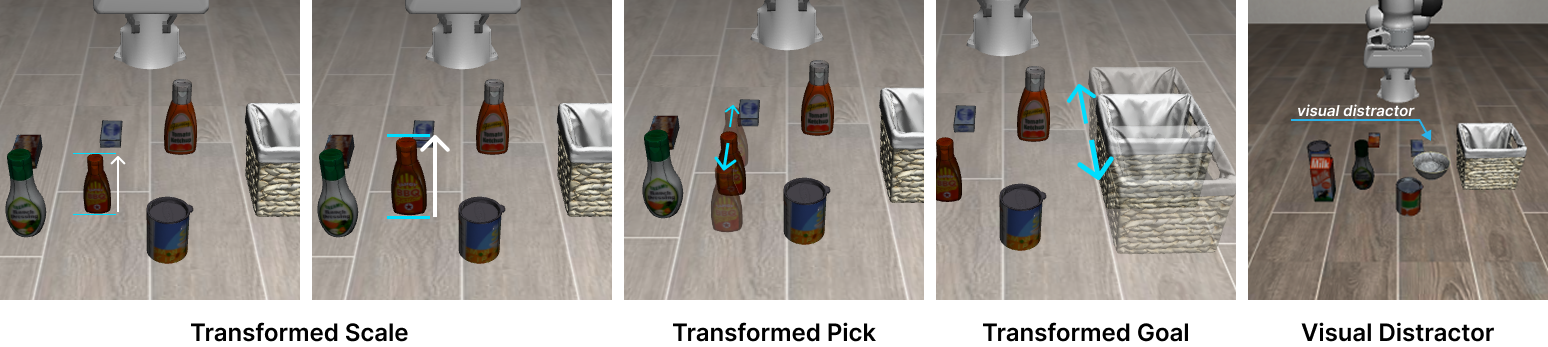}
    \caption{Visualization of axes constructed in Libero-Object suite. Three axes displayed are (1) Transformed scale of the object (2) Perturbed pick location of the object (3) Perturbed goal location of the basket. We visualization of additional test on visual distractors.}
    \label{fig:libero_object_sim}
\end{figure*}

\textbf{LIBERO Experiments.} LIBERO-Object ~\citep{liu2023libero} provides 10 pick-and-place tasks and serves as the primary finetuning benchmark for the VLAs we evaluate. We denote the ID task-space configuration as $\tau_0$ and generate physically feasible perturbations along three axes indexed by $k \in \{-3,-2,-1,0,1,2,3\}$. Larger values of $|k|$ correspond to increasing deviation from training support, producing the nested structure $\tau_0 \subset \tau_{\pm1} \subset \tau_{\pm2} \subset \tau_{\pm3}$. We instantiate the following three axes: (1) \textit{Pick object translation axis.} The pick object is translated along the $y$-axis with step size $2.5$ cm. (2) \textit{Goal location translation axis.}
Similarly, the basket goal location is translated with step size $3.67$ cm. (3) \textit{Scale perturbation axis.}
Objects are scaled uniformly with a step size of $6.67\%$. Step size values are generated to ensure tasks stay within limits of workspace, remain functional and graspable. We validate four SOTA VLA models: $\pi_{0.5}$ \cite{black2025pi}, OpenVLA-OFT \cite{kim2025fine}, UniVLA \cite{wang2025unified} and RynnVLA \cite{cen2025rynnvla} using provided pre-trained checkpoints on LIBERO. The results of these experiments are shown in Figure \ref{fig:libero_object_sim}. For each perturbed task we conducted 50 evaluation rollouts in simulation. All models achieve near-ceiling success at $\tau_0$, making standard in-distribution evaluation uninformative for distinguishing their generalization capability. Along spatial axes, however, clear performance degradation emerges as $|k|$ increases (Figure \ref{fig:libero_results}).

\textbf{RoboCasa Experiments.} 
LIBERO's limited visual and task diversity raises the question of whether the observed failures are artifacts of a narrow benchmark. We scale our methodology to RoboCasa~\citep{nasiriany2026robocasa365} and construct 3 inductive axes: spatial (pick and goal), task horizon length and clutter. We cover these progressions in detail in Appendix~\ref{app:robocasa}.

\textbf{Result: Inductive generalization failures persist across benchmarks, scales,
and data regimes.} Across both LIBERO-Object and RoboCasa, all models achieve strong ID performance, yet performance degrades consistently as tasks move
 along structured axes (Figure~\ref{fig:libero_results},
Figure~\ref{fig:robocasaoverall}). This pattern holds even
in RoboCasa, where policies (Isaac-GR00T~\cite{bjorck2025gr00t} and $\pi_{0.5}$~\cite{black2025pi}) are exposed to substantially greater visual diversity
and domain randomization during training, and where finetuning on additional
MimicGen-generated ID data yields no improvement in inductive
generalization. Tests on visual distractors, which perturb only the visual
input without affecting task dynamics, produce no
measurable change in policy performance on their own. By composing this with the pick axis, the degradation is attributable to the model's inability to handle novel object poses, ruling out visual confusion as a confounding factor. In addition, the uniform and symmetric nature of evaluated objects in LIBERO exhibits an implicit scale equivariance; this finding is consistent with real-robot results in Section~\ref{sec:realrobotexp}. This result demonstrates that task environments can encode certain structural invariances while missing other inductive axes entirely. This result signals that data scale, visual diversity, nor task design alone closes the inductive
generalization gap. These results call for future work that studies the relationship between data and algorithmic design that generalizes inductively instead of loosely ID.

\subsection{Real Robot Experiments}
\label{sec:realrobotexp}
\begin{wrapfigure}{r}{0.45\textwidth}
    \centering
    \captionof{table}{Success rate of structured difficulty progressions on the real robot with the \(\pi_{0.5}\) policy. Increasing $\tau_k$ corresponds to increasing deviation from the training distribution. The worst drop is in red.}
    \label{tab:real_robot_progressions}
    \setlength{\tabcolsep}{3pt}
    \renewcommand{\arraystretch}{1.05}
    \resizebox{\linewidth}{!}{
    \begin{tabular}{lccccc}
        \toprule
        \textbf{Axis} & $\bm{\tau_0}$\,(ID) & $\bm{\tau_1}$\,(OOD) & $\bm{\tau_2}$\,(OOD) & $\bm{\tau_3}$\,(OOD) & $\bm{\tau_4}$\,(OOD)\\
        \midrule
        Stack & 70\% & 30\% & \drop{0\%} & --   & --   \\
        Scale & 90\% & 100\% & 90\%      & 100\% & \drop{70\%} \\
        \bottomrule
    \end{tabular}}

    \vspace{0.6em}

    % --- Figures below ---
    \begin{subfigure}{\linewidth}
        \centering
        \includegraphics[width=\linewidth]{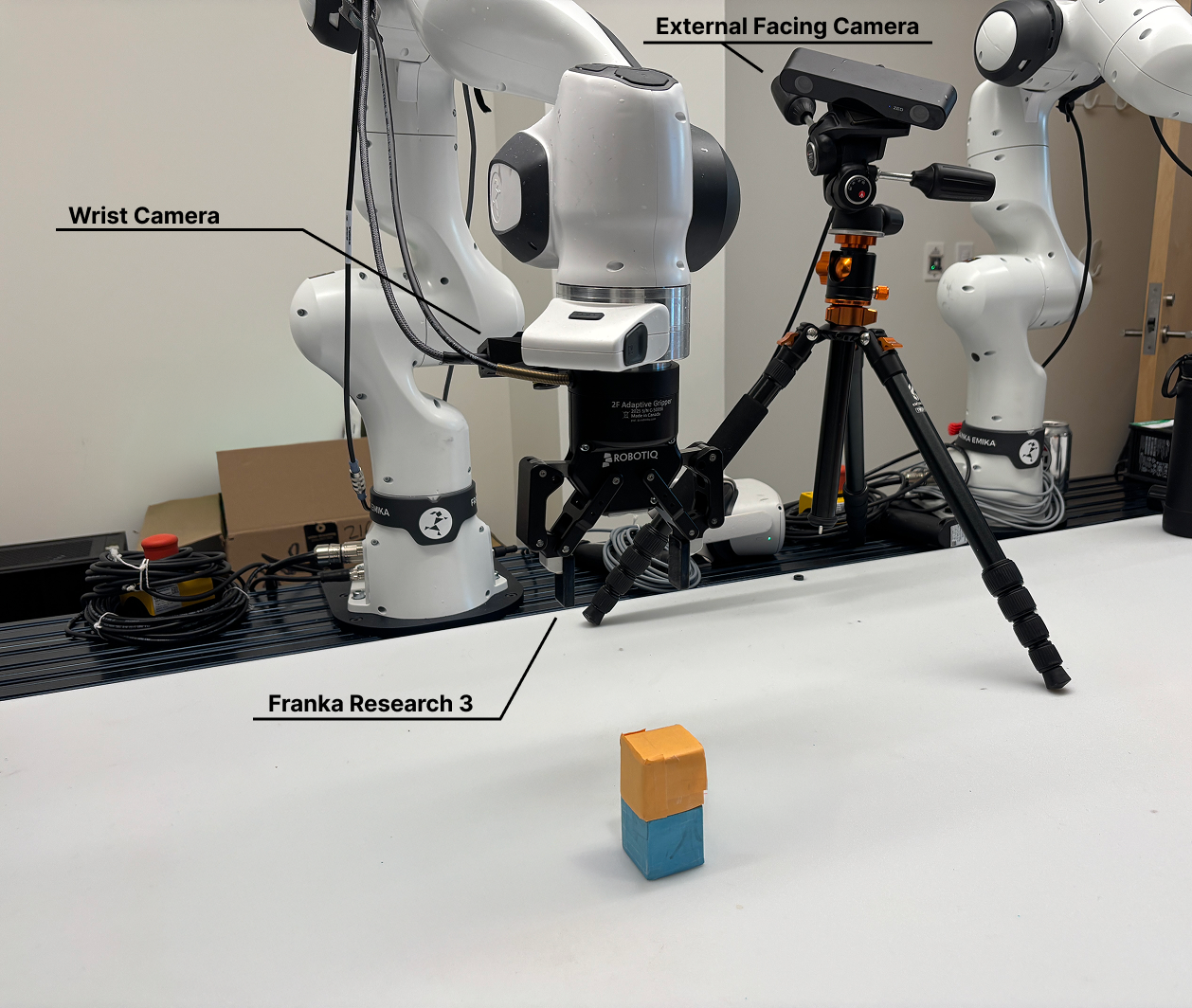}
        \label{fig:real-world-exp-setup}
    \end{subfigure}

    \vspace{-0.3em}

    \caption{Real-world setup and evaluation objects.}
    \label{fig:real_world}

\end{wrapfigure}

To confirm that inductive axes are constructable in real, we perform inference of $\pi_{0.5}$, (finetuned on the DROID dataset \cite{khazatsky2024droid}) on a Franka Research 3 robot with the DROID setup (Figure \ref{fig:real_world}). In Table \ref{tab:real_robot_progressions} we construct two tasks, drawn from the DROID dataset: (1) `Stack' - ``Place the orange block on the blue block" and (2) `Scale' - ``Pick up the cup and put it in the black bowl". The objects used for the real-world experiment are shown in Figure \ref{fig:realworldobjectsvis}. `Stack' undergoes increasing z-height (incremented by 0.25 inches) to simulate the notion of stacking an additional cube. `Scale' induces a progression in size of the graspable bowl. Policy's results to these perturbations are shown in  Table \ref{tab:real_robot_progressions}. Lighting, background, and camera pose are fixed. First, we ensured that the policy completed the base task $\tau_0$ with a high success rate then we perturbed the policy along the underlying axis until failure. For cube stacking we find that increasing z-height quickly leads to failure as the policy cannot output actions past a narrow region of height, causing toppling of the stack. Policy remained robust on cups of different scale, due to the object symmetry, aligning with results found in simulation. Therefore, we display an example of how to construct and execute inductive generalization testing in real robot experiments. While this provides insights on real world performance, we believe simulation remains the best option for testing inductive generalization ability at scale.
\vspace{-0.5em}
\section{Conclusion and Limitations}
\label{sec:conclusion}
\vspace{-0.5em}
As robotic policies scale in data and model capacity, the standard of measuring generalization must evolve to measure the ability to act beyond observed experience. Our experiments show that evaluated VLAs fail along structured axes and supplementation of additional ID data does not help. Our results point toward researching learning invariant behavior rules either through architectures, action heads or engineering inductive biases. We emphasize, our \textit{inductive generalization} method is not intended to replace existing notions of generalization; e.g, sim-to-real transfer, or domain randomization. Instead, we hope this framework provides a foundation for systematically studying how architectural choices, training objectives, and data scaling influence the emergence of rule-level abstractions in visuomotor policies, and ultimately contributes to building robotic systems that understand tasks rather than merely fit them.

\textbf{Limitations.} 
Our methodology relies on constructing controlled, low-dimensional axes of variation (e.g., spatial, horizon-length, recursion) that isolate specific factors of task difficulty. While this enables diagnostic analysis of generalization behavior, it necessarily simplifies the full complexity of robotic environments, where variations occur jointly across many correlated factors. As a result, performing well along one controlled axis does not mean a model will remain robust when many factors change at once or in less structured ways. In addition, the choice of axes reflects designer assumptions about which task attributes are meaningful. Although we focus on physically interpretable properties such as geometry and spatial layout, other sources of variation may induce different generalization patterns that our current axes do not capture. In addition, while our evaluation is grounded in tabletop settings, the proposed framework is not domain-specific and can be readily adapted to other settings, such as mobile and dexterous manipulation, thus leaving ample room for future work.
%===============================================================================

\clearpage
% The acknowledgments are automatically included only in the final and preprint versions of the paper.
% \acknowledgments{If a paper is accepted, the final camera-ready version will (and probably should) include acknowledgments. All acknowledgments go at the end of the paper, including thanks to reviewers who gave useful comments, to colleagues who contributed to the ideas, and to funding agencies and corporate sponsors that provided financial support.}

%===============================================================================

\bibliography{example}  % .bib

\clearpage
\appendix

\section{BlockEnv Experiments Extended}
\label{app:blockenv}

We design a Block Environment that follows an inductive structure. In other words, the task admits a heuristic that generalizes to all levels $k$, yet no training distribution at level $k$ provides coverage for the action at level $k{+}1$. 

\begin{figure}[h]
    \centering
    \includegraphics[width=0.75\linewidth]{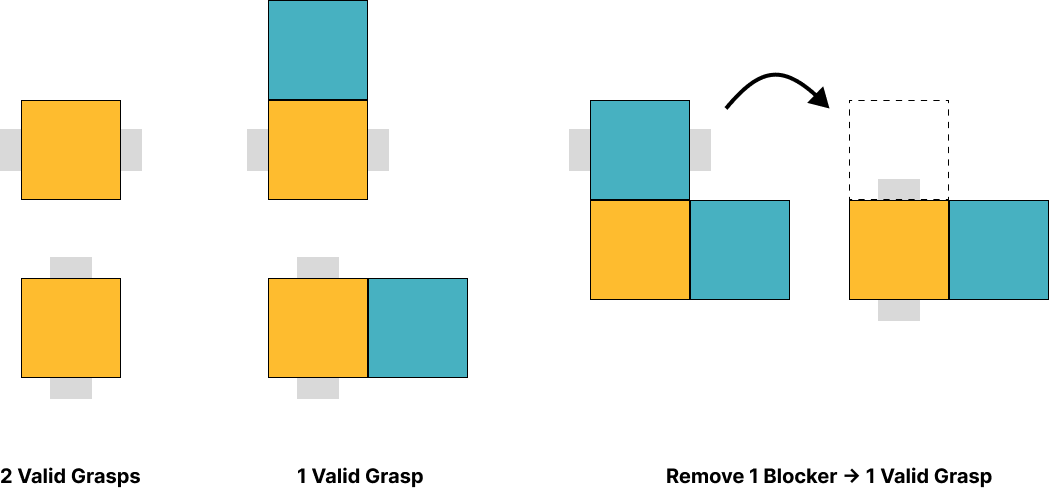}
    \caption{Overview of grasp validity for the Block Environment task space. Each level introduces distractors which require removal in order to reveal a valid grasp of the goal object (yellow).}
    \label{fig:placeholder}
\end{figure}

\textbf{Task definition.} At level $k=0$ the scene contains only the goal block. At level $k \geq 1$, the environment places $2k$ distractor blocks by constructing two symmetric chains of length $k$ that emanate from the goal position: a forward chain and a flipped chain, each grown one cell at a time by alternating row- and column-offsets according to the recursion depth. Concretely, the $i$-th block in the forward chain is offset from the $i{-}1$-th block by one step downward if $i$ is even and one step to the left if $i$ is odd (and vice-versa for the flipped chain). The resulting structure is then translated to a uniformly-random position within the grid, so neither the spatial location nor the absolute coordinates of any block are predictable across episodes.

\textbf{Observation and action space.} The agent observes a flat $G \times G$ occupancy grid (values ${0, 1, 2}$ for empty, distractor, and goal respectively). An action is a triple $(r, c, \theta)$ where $(r, c)$ selects a grid cell and $\theta \in {0, 1}$ specifies a top/bottom or left/right grasp orientation. A grasp attempt succeeds only when the cell is non-empty and neither adjacent cell along the chosen approach axis is occupied.

\textbf{Inductive structure.} The heuristic that solves all levels is: repeatedly identify and remove any distractor block that currently has a free approach axis, until the goal block becomes graspable. This rule is expressible at every level $k$, yet an agent trained exclusively on level-$k$ demonstrations never observes the new outermost distractor configuration introduced at level $k{+}1$. The outermost distractors at level $k{+}1$ occupy positions that are strictly outside the convex hull of any level-$k$ arrangement, ensuring zero state-coverage overlap for the first required action of the harder task. This makes the Block Environment a controlled testbed for measuring whether a learning-based algorithm can extrapolate an acquired a skill which transfers to unseen compositional depth.

\subsection{Baselines}

We compare against the following baselines, all trained via behavior cloning on oracle demonstrations from levels $k \in \{0, 1, 2\}$.

\textbf{RNN.} A goal-conditioned 2-layer LSTM policy.
At each step $t$ the flattened grid observation $o_t \in \mathbb{R}^{G^2}$ is concatenated with a normalized goal scalar $g = k/k_{\max}$ and passed through an MLP encoder before the LSTM, which maintains a hidden state across the episode. Three independent linear heads predict logits over rows, columns, and orientations respectively. While normalizing the goal isn't appropriate for true inductive generalization, to provide this baseline with the best shot we allowed normalization.

\textbf{Transformer.} Identical input encoding to the RNN, but the LSTM is replaced by a causal Transformer Encoder with learned absolute positional embeddings.
The full observation history is re-processed at every step, with the action read from the last-position output.

\textbf{Transformer + RoPE.}
The same causal Transformer architecture, but with learned absolute positional embeddings replaced by Rotary Position Embeddings (RoPE)~\cite{su2024roformer}, which encode relative temporal offsets directly inside each attention head and add no additional parameters.
Pre-norm (LayerNorm before each sub-layer) is used for training stability.

\textbf{Transformer + Goal/Action.}
A further augmentation of Transformer + RoPE introducing two structural inductive biases for compositional tasks:
(i) the initial grid $o_0$ is prepended as a special goal token so the model can always attend back to the target configuration; and
(ii) the previous action $(r_{t-1}, c_{t-1}, \theta_{t-1})$, normalized to $[0,1]^3$, is appended as an explicit feature to each observation token, giving the model direct memory of its own decision sequence.

\textbf{Ind (Ours).}
Built on Transformer + RoPE with two targeted additions motivated by the grasp-legality structure of the task.
First, an auxiliary \emph{legality head} produces per-cell, per-orientation logits predicting whether each grasp is unobstructed; the target is computed analytically from the current observation (no oracle information) and is consistent across all levels.
Second, a factored $(r,c,\theta)$ action head jointly scores cells and orientations; at inference, unoccupied cells are hard-masked to $-\infty$ and the legality head additively gates the pick logits, structurally preventing the policy from attempting physically blocked grasps.

\textbf{MAML.}
Model-Agnostic Meta-Learning~\cite{finn2017model} over the three training levels treated as separate tasks. The meta-objective finds an initialization $\theta$ such that a \emph{single} inner gradient step on a small support set of demonstrations from each level yields a policy that solves that level.
At evaluation on a held-out level, the model is adapted using a few-shot support set before rollout (FOMAML variant; inner $\alpha = 0.05$, one inner step).

\textbf{TRM.}
Tiny Recursive Model~\cite{jolicoeur2025less}, a deep-supervision architecture that frames each single-step grid observation as an independent puzzle.
The model maintains two latent tensors, $y$ (current answer) and $z$ (reasoning state), and iterates
$z \leftarrow f(z,\, x + y)$ for $L$ cycles, then
$y \leftarrow f(y,\, z)$ once, repeating this for $T{-}1$ outer cycles without gradients followed by one cycle with gradients.
An adaptive computation-time (ACT) halting mechanism with Q-learning decides how many outer cycles to execute at runtime.
The action is decoded from the output grid by reading the cell marked with an action token.

\textbf{Oracle.}
The scripted oracle implements the inductive heuristic directly: it selects one arm of the distractor chain uniformly at random, then iteratively removes the tip-most block that has a free approach axis, working inward until the goal block is exposed and graspable.
This procedure is sound and complete at every level $k$ by construction, and is used solely as a \emph{demonstrator} to generate training data; it is never shown to the learned policies at test time.

\section{RoboCasa Experiments Extended}
\label{app:robocasa}

\textbf{Original Policy Performance.} RoboCasa classifies tasks into atomic and composite tasks. For the purposes of our study, we solely focus on atomic tasks. In order to study a policy trained only on atomic tasks, we fine-tuned Isaac-GR00T and Pi0.5 solely on atomic data provided in RoboCasa. In addition, we fine-tuned an additional checkpoint for Isaac-GR00T on additional provided MimicGen data, totaling approximately 1600 collected hours.

\textbf{Black-Box Inductive Generalization.}
Definition~\ref{def:structured_difficulty} assumes the base task distribution
$\mathcal{T}_0$ is known by construction, which we could satisfy in LIBERO-Object. However, generally speaking for VLAs and subsequent RoboCasa experiments, the training distribution is not fully observable, making it impossible to certify that any
configuration is provably ID. We therefore distinguish two evaluation regimes: (1) \emph{Vanilla IndGen.} The training distribution is controlled;
    $\mathcal{T}_0$ is defined by construction and Condition~2 of Definition~\ref{def:inductive_gen} holds analytically. This regime applies to our LIBERO and BlockEnv experiments. (2) \emph{Black-box IndGen.} The training distribution is unknown or excessively broad (e.g., a VLA pretrained on internet-scale robot data).
    Here, $\mathcal{T}_0$ cannot be certified distributionally and is instead defined empirically.

\begin{definitionbox}[Black-Box Reference Level, label=def:blackbox_reference]
Let $\Pi$ be a reference policy class and let
$\{\tau^{(j)}\}_{j=1}^{J}$ be a set of canonical task configurations.
The \emph{black-box reference level} $\mathcal{T}_0$ is defined as
\[
    \mathcal{T}_0 \;=\; \operatorname*{arg\,max}_{\tau^{(j)}}\;
    \mathbb{E}_{\pi \sim \Pi}\bigl[S(\pi,\, \tau^{(j)})\bigr],
\]
i.e., the configuration that maximizes expected success rate over the
reference policy class under standard evaluation conditions.
\end{definitionbox}

\noindent
In the black-box regime the strict out-of-support claim in
Condition~2 is replaced by an empirical ordering requirement. In this setting the
difficulty progression is \emph{validated post hoc} by observing that
$S(\pi, \mathcal{T}_k)$ decreases monotonically as $k$ increases. Conditions~1 and~3 of Definition~\ref{def:structured_difficulty}
(semantics preservation and single-axis control) remain
analytically enforceable in both regimes and are satisfied by construction in all experiments that follow.

\subsection{Extended Inductive Generalization Experiment Details}

\begin{figure}[h]
    \centering
    \includegraphics[width=\linewidth]{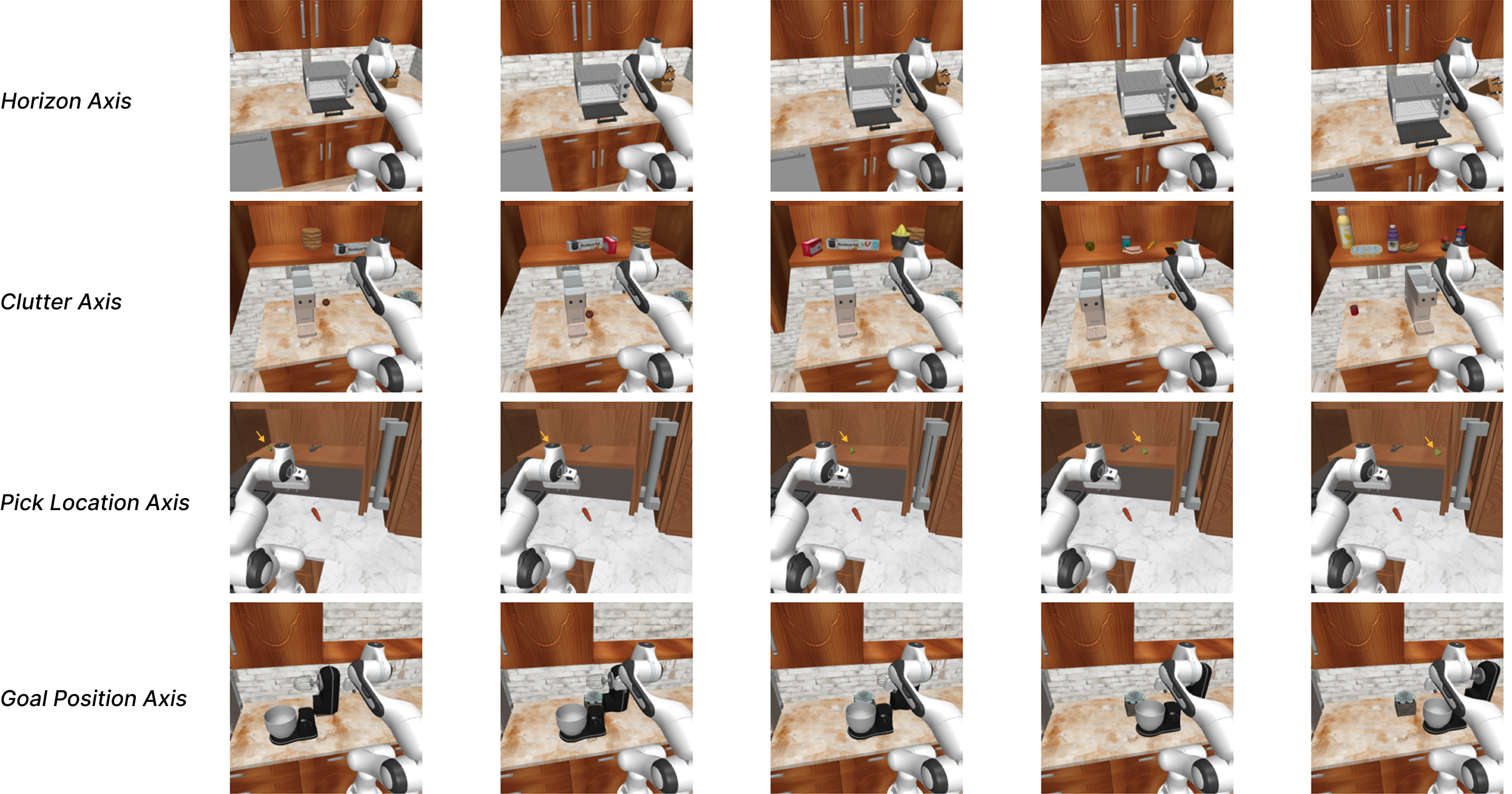}
    \caption{Sample visualization of perturbed task axes.}
    \label{fig:robot_location_x_vis}
\end{figure}

 For each task we apply three axes of induction: (1) Spatial, (2) Task Horizon, and (3) Clutter. We note that the original data collected in RoboCasa was collected quite broadly, so for our Pick and Spatial axes we do not have a sole in-distribution $\tau_0$ as the Clutter axis does. However, we do guarantee that for increasing $\tau$ tasks become increasingly harder as shown empirically by decreasing success rates in Figure~\ref{fig:robocasaoverall}. In addition, since RoboCasa includes numerous atomic tasks, some with better performance than others, we ensured to pick atomic tasks that policies performed well in-distribution. We include a sample visualization of perturbations along axes in Figure~\ref{fig:robot_location_x_vis}.

\textbf{Spatial Axis Results.} In this task, we perturb the goal and pick object locations in order to simulate an object-centric inductive axis.For the pick setting, we perturb objects along the x-axis in the following atomic tasks: CheesyBread, CoffeeSetupMug, PickPlaceCabinetToCounter, PickPlaceCounterToBlender, PickPlaceCounterToCabinet, PickPlaceCounterToDrawer, PickPlaceCounterToSink, PickPlaceCounterToStandMixer, and PickPlaceSinkToCounter. The perturbation amount takes the valid sampling region and divides it by number of progressions along the axis (e.g. 5). This ensures that objects are always valid and graspable. For the goal setting, we perturb fixtures along the x-axis in the following atomic tasks: CloseStandMixerHead, OpenStandMixerHead, TurnOnElectricKettle, OpenElectricKettleLid, CloseElectricKettleLid, and TurnOnBlender. Fixtures here are perturbed 12.5cm per step.

\textbf{Horizon Task Axis Results.} In this task, we perturb the robot base along the y-axis on 6 atomics tasks in order to simulate the effect of increasing task horizon length. The atomic tasks are the following: PickPlaceFridgeDrawerToShelf, CloseToasterOvenDoor, PickPlaceCounterToDrawer, PickPlaceCounterToSink, PickPlaceCounterToCabinet, and PickPlaceCounterToBlender. By moving the robot further away from original task settings (12.5 cm per step), it requires more time for the robot to reach and manipulate objects. Thus, we analyze how changing the effective time horizon of task impacts the policy's performance.

\textbf{Clutter Axis Results.} In this task PickPlaceCabinetToCounter, we add additional random objects inside a cabinet to simulate clutter. $\tau_0$ represents the original task setting, and increasing $\tau$ corresponds to an additional item added to the cabinet. A cluttered cabinet requires the policy to reason about how to navigate around the clutter, and potentially manipulate the clutter to make access to the goal object feasible. Policy's have been trained on manipulating all objects individually in pre-training. However, despite this performance is not maintained along this inductive axis. This axis was introduced in particular because clutter is inherently NP-Hard and traditionally solved using motion planning techniques. However, its unclear whether end-to-end policies will eventually display emergent techniques to solve tasks such as these. 

\begin{figure}[h]
    \centering
    \includegraphics[width=0.5\linewidth]{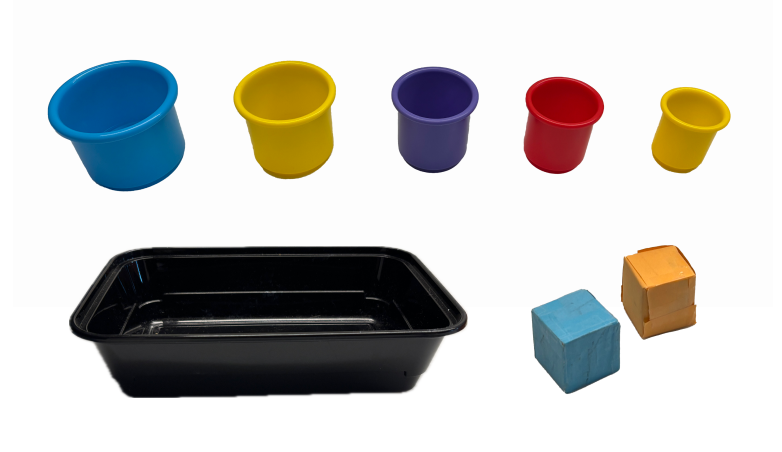}
    \caption{Objects used in real robot experiments.}
    \label{fig:realworldobjectsvis}
\end{figure}

\section{Robotic Manipulation Benchmark Evaluation Methods}
\label{app:benchmark}

\textbf{Open X-Embodiment}~\cite{open_x_embodiment_rt_x_2023}: 
Open X-Embodiment is focused on zero-shot cross-embodiment generalization, primarily addressing the ability of policies to adapt to tasks on unseen robotic platforms. In contrast to many traditional benchmarks and datasets, Open X-Embodiment utilizes an aggregate dataset (consisting of data from over 60 individual datasets) spanning multiple real-world platforms (Franka, UR5, etc.)~\cite{open_x_embodiment_rt_x_2023}. While this data is useful in pre-training it still has yet to be shown how the quantity and diversity of such data assists in generalization on OOD task spaces. 

\textbf{ManiSkill}~\cite{tao2024maniskill3}: 
ManiSkill also focuses on zero-shot generalization, but specifically intra-category generalization. For example, if a category of tasks $C$ consists of instances $\{c_1, \dots, c_N\}$, the policy $\pi$ is trained on $C_{train} \subset C$ and tested on $C_{test} \subset C$ where $C_{train} \cap C_{test} = \emptyset$. Further, it also leverages a high-fidelity physics engine that allows for complex rigid-body articulation and uses kinematic success constraints, utilizing factors such as joint angles of articulated objects and constraining contact forces to ensure the policy is physically feasible and valid.

\textbf{RoboSuite}~\cite{zhu2020robosuite}:
RoboSuite evaluates single task performance using cumulative reward $R = \sum_{t=0}^T \gamma^t r_t$, focusing on sample efficiency (how many environment steps are needed to achieve 90\% success). Single task performance involves evaluating the policy on tasks that are directly present in the training distribution, albeit with some variance to prevent rote memorization. In evaluation, this environment randomization follows a distribution $p(\xi)$, within which certain physics parameters $\xi$ (mass $m$, friction $\mu$, damping $d$, etc.) along with visual parameters (camera pose, lighting, etc.) are heavily perturbed. Evaluation then measures the expected return under these perturbations, which is given by $\mathbb{E}_{\xi \sim p(\xi)} [R(\pi, \xi)]$~\cite{tobin2017domain}.

\textbf{LIBERO}~\cite{liu2023libero}:
LIBERO takes a unique approach in evaluating continual learning, i.e., how well a policy learns a series of tasks $\mathcal{T}_1, \mathcal{T}_2, \dots, \mathcal{T}_N$ without overwriting previous knowledge. It is built on a RoboSuite backend, but on top of that, there is a structured framework consisting of 130 sequential tasks divided by semantic terms. LIBERO measures not only the average performance $A_N$ (overall competence), but also backward transfer $F_N$, how performance on old tasks degrades after learning new ones, and forward transfer $FW_N$, whether learning earlier tasks accelerated the learning of the current task.

\textbf{LIBERO-PRO}~\cite{zhou2025liberoprorobustfairevaluation} and \textbf{LIBERO-Plus} \cite{fei2025liberoplusindepthrobustnessanalysis} extend LIBERO by broadening the set of perturbations in object color, position, layout, and camera viewpoint. While these efforts align with our motivation to assess generalization in structured ways, they primarily diversify perturbation types and do not identify where axes lie within existing training distributions and their separation from test-time evaluation.

\textbf{RLBench}~\cite{james2020rlbench}:
Unlike zero-shot benchmarks that utilize an entirely unseen test set, RLBench uses few-shot meta-learning where the model is trained on a distribution of meta-training tasks $p(\mathcal{T})$. During evaluation, it is given an unseen task $\mathcal{T}_{new}$ and a small set of $K$ demonstrations ($K \ll \lvert p(\mathcal{T}) \rvert$). Success is then calculated on a query set $\mathcal{T}_Q \subset \mathcal{T}_{new}$ after applying $N$ gradient update steps based only on the $K$ demonstrations.

\section{Further Discussion}
\label{sec:discussion}

\textbf{Why do we need to construct axes of generalization? Shouldn’t data we train on inherently contain structures?} Yes, as there is always an underlying non-random data-generating process, the data are samples that should hint at the underlying structure that generated them. However, for the limited data setting of robotics, it is helpful to construct explicit, human-readable axes where structures are known in order to test that models are learning basic underlying structures. As we show in our experiments, even with simple tests, this is not the case at the moment. Of course in the future, removing the need for human-readable axes and allowing models to learn unknown biases is the goal and will be a focus in future works as our methods improve.

\textbf{What about SE(3) Equivariance or other Inductive Biases, isn't this already researched?} Our argument is not that inductive biases are unimportant or underexplored, but rather that they are traditionally introduced surgically to capture a previously identified structure.  Our aim is instead to incentivize the creation of learning paradigms that uncover structure from data, and make use of these learned behaviors beyond a specific known condition. Our argument is intentionally model-agnostic because a central hypothesis in the large Visuomotor Policy and Foundation Model literature is that scale will give rise to generalizable learned representations \cite{black2025pi, kim2024openvla, li2025dsvla, zhen20243d, hu2023toward, open_x_embodiment_rt_x_2023}. If this is true, it would manifest directly as maintained performance across structured axes. Our framework therefore creates the conditions under which the presence or absence of learned (or injected) inductive biases becomes observable. By reliably surfacing these failures it is the first step towards informed architectural or training choices.

\textbf{Infinite Scenarios.} The proposed framework does not require exhaustive coverage of proposed axes. Instead, one only needs to sample discretely along the axis in order to diagnose whether a model has learned the underlying axis rule. While denser sampling may reveal higher fidelity of performance curves there are vanishing returns. However, the quantity of data required to uncover such priors remains unknown but is countably small for human learners. 
\end{document}